\documentclass[journal=accacs,manuscript=article]{achemso}
\usepackage{amssymb}
\usepackage{amsmath,bm}
\usepackage{graphicx,color}
\usepackage[version=3]{mhchem} 
\usepackage{xcolor}
\usepackage{soul}
\sethlcolor{yellow}
\usepackage{multirow}

\newcommand{\orcid}[1]{\href{https://orcid.org/#1}{\textcolor[HTML]{A6CE39}{\aiOrcid}}}

\author{Joyjit Chattoraj}
\email{joyjit_chattoraj@ihpc.a-star.edu.sg}
\affiliation{Computing and Intelligence, Institute of High Performance Computing, Agency for Science Technology and Research, Singapore 138632, Republic of Singapore}
\author{Brahim Hamadicharef}
\affiliation{Computing and Intelligence, Institute of High Performance Computing, Agency for Science Technology and Research, Singapore 138632, Republic of Singapore}
\author{Teo Shi Chang}
\affiliation{Department of Catalysis and Green Process Engineering, Institute of Sustainability for Chemicals, Energy and Environment, Agency for Science Technology and Research, Singapore 627833, Republic of Singapore} 
\author{Yingzhi Zeng}
\affiliation{Materials Science and Chemistry, Institute of High Performance Computing, Agency for Science Technology and Research, Singapore 138632, Republic of Singapore}
\author{Chee Kok Poh}
\affiliation{Department of Catalysis and Green Process Engineering, Institute of Sustainability for Chemicals, Energy and Environment, Agency for Science Technology and Research, Singapore 627833, Republic of Singapore}
\author{Luwei Chen}
\affiliation{Department of Catalysis and Green Process Engineering, Institute of Sustainability for Chemicals, Energy and Environment, Agency for Science Technology and Research, Singapore 627833, Republic of Singapore}
\author{Teck Leong Tan}
\affiliation{Materials Science and Chemistry, Institute of High Performance Computing, Agency for Science Technology and Research, Singapore 138632, Republic of Singapore}

\title{AceWGS: An LLM-Aided Framework to Accelerate Catalyst Design for Water-Gas Shift Reactions} 
\keywords{Water-Gas Shift (WGS), Catalyst Design, Inverse Modelling, Large Language Model (LLM), Generative AI}
\begin{document}

\begin{abstract}
While the Water-Gas Shift (WGS) reaction plays a crucial role in hydrogen production for fuel cells, finding suitable catalysts to achieve high yields for low-temperature WGS reactions remains a persistent challenge.
Artificial Intelligence (AI) has shown promise in accelerating catalyst design by exploring vast candidate spaces, however, two key gaps limit its effectiveness. First, AI models primarily train on numerical data, which fail to capture essential text-based information, such as catalyst synthesis methods. Second, the cross-disciplinary nature of catalyst design requires seamless collaboration between AI, theory, experiments, and numerical simulations, often leading to communication barriers. 
To address these gaps, we present AceWGS, a Large Language Models (LLMs)-aided framework to streamline WGS catalyst design. AceWGS interacts with researchers through natural language, answering queries based on four features: (i) answering general queries, (ii) extracting information about the database comprising WGS-related journal articles, (iii) comprehending the context described in these articles, and (iv) identifying catalyst candidates using our proposed AI inverse model. We presented a practical case study demonstrating how AceWGS can accelerate the catalyst design process. 
AceWGS, built with open-source tools, offers an adjustable framework that researchers can readily adapt for a range of AI-accelerated catalyst design applications, supporting seamless integration across cross-disciplinary studies.
\end{abstract}

\section{Introduction}
\label{sec:intro}
The increasing global energy demand and reliance on carbon-based fuels significantly contribute to environmental pollution. Hydrogen, produced from renewable sources, offers a sustainable, carbon-free alternative and plays a key role in decarbonizing the global energy system, especially through fuel cell technologies. 
Hydrogen used in fuel cells is commonly produced through hydrocarbon reforming processes, which often yield undesirable by-products. For example, in proton exchange membrane fuel cells, carbon monoxide (CO) is a common contaminant in the hydrogen fuel, arising as a by-product of production methods such as steam methane reforming.
This CO must be entirely removed to protect the anode catalyst. 
The Water-Gas Shift (WGS) reaction, a catalytic process between CO and H$_2$O to produce H$_2$ and CO$_2$, is seen as the solution. For fuel cells, WGS catalysts need to be highly stable, active, and able to function without special pretreatment or regeneration to achieve maximum CO conversion at low temperatures. Noble metal catalysts are potential candidates for such applications~\cite{Oensan2007, Park2009, Ebrahimi2020, ZhouLiu2023, ShuiJZZLH2023}.

Artificial Intelligence (AI) has become an increasingly appealing approach for catalyst design, as advanced AI models can uncover complex relationships between numerous variables, enabling them to explore and exploit the vast design space~\cite{Toyao2020, Benavides2024}.
In the context of AI research on the WGS reaction, Odaba\c{s}i et al.~\cite{Odabasi2014} conducted the first comprehensive study. They developed a database using data mining techniques, comprising 4,360 experimental data points and 81 features, including catalyst compositions, preparation methods, reaction conditions, and CO conversion. These data were extracted from 84 research articles published between 2002 and 2012.
They also developed AI models to predict CO conversion based on 80 other features. Several subsequent studies have utilized this database, proposing various AI models and techniques to more effectively capture the correlation between catalyst characteristics and CO conversion~\cite{Avsar2017, Cavalcanti2019, Suzuki2019, Smith2020, Garcia2023, golder2024machine}.

In the recent past, the authors of this article developed a theory-guided AI model, training it on the same database while incorporating thermodynamic equilibrium constraints through a custom loss function. They demonstrated that their AI model strictly adhered to the thermodynamic equilibrium principle, leading to more accurate and robust predictions~\cite{Chattoraj2022}. 
Following this work, the authors compiled a new WGS database by extracting data from 82 articles published between 2013 and 2021. This effort produced 8,908 individual records with 99 features, covering 10 different base metals, 27 supports, 16 promoters, 32 preparation methods, 13 reaction conditions, and carbon monoxide conversion percentages.
Furthermore, they developed an inverse model that integrates the theory-guided AI model with a particle swarm optimization method. This inverse model can explore and utilize the new database to identify suitable catalysts for low-temperature WGS reactions based on the design constraints set by the researchers~\cite{Chattoraj2023}.

The primary limitation of the AI models that are mentioned above is their reliance solely on numerical data. For example, they reduce complex catalyst preparation methods, such as wet impregnation, to simple categorical variables in a {\it yes} or {\it no} format, whereas, catalyst synthesis involves a series of detailed, multi-step processes. Therefore, when an inverse model predicts a catalyst design, it cannot provide the corresponding step-by-step synthesis procedure, which is crucial for practical implementation. This limitation underscores the need to utilize the textual content of relevant research articles.

Another limitation of current AI-accelerated catalyst design is the necessity for cross-disciplinary collaboration. Establishing a comprehensive AI pipeline, beginning with database preparation, data cleaning, feature selection, and extraction, and extending to training AI models, including inverse models, and validating predictions through simulations and experiments, often requires the involvement of multiple researchers with diverse expertise. This complexity can hinder the efficiency of the research process, as it demands effective communication and coordination among specialists in fields such as chemistry, materials science, computer science, and engineering. Consequently, the multifaceted nature of AI-driven catalyst design can create challenges in workflow integration and knowledge transfer, potentially slowing the advancement of innovative catalyst designs.

Large Language Models (LLMs), such as ChatGPT, could offer significant advantages in addressing existing limitations in AI-accelerated catalyst design. These models analyze vast amounts of text data, utilizing deep learning techniques including neural networks and transformers to comprehend context, predict words, and generate human-like responses. By learning complex patterns from the data, LLMs produce coherent and contextually relevant answers during conversations. The integration of Retrieval-Augmented Generation (RAG) enhances this capability by connecting LLMs with external knowledge bases, enabling access to up-to-date information for generating reliable outputs. 
 The synergy of LLMs and RAG, referred to as LLM-RAG, has profound implications for scientific research. LLM-RAG systems can provide accurate and dependable answers to scientific inquiries by leveraging the reasoning capabilities of LLMs alongside real-time information retrieval. RAG's ability to manage long contexts and its high interpretability make it particularly suitable for complex, integrative, or summary questions that necessitate processing large volumes of material~\cite{fan2024survey, kumar2024can,si2024can}.

Kim et al.~\cite{kim2024large} fine-tuned LLMs to predict the synthesizability of inorganic compounds and select synthesis precursors, demonstrating promising performance compared to specialized AI models.
Wang et al.~\cite{wang2024catalm} introduced a large language model specifically tailored for the domain of electrocatalytic materials, demonstrating its potential to enhance human-AI collaboration in catalyst knowledge exploration and design.
Bran et al~\cite{bran2024augmenting} developed ChemCrow, a chemistry-focused LLM-RAG system using GPT-4 and 18 expert-designed tools. The system enhances organic synthesis, drug discovery, and materials design performance, effectively automating chemical processes and bridging the gap between experimental and computational chemistry.

In this article, we introduce an LLM-RAG framework, AceWGS, that aims to accelerate the design of noble metal catalysts for the WGS reaction. 
AceWGS maximizes the utilization of both text and numerical data from the research articles on WGS experiments involving noble metals as catalysts, while also demonstrating the advantages of combining traditional AI models with LLMs.
The framework comprises four key features: (i) a generic querying tool that addresses researcher queries related to WGS reactions and AI methods, (ii) an extraction tool that retrieves information from a local WGS database of 82 research articles, (iii) a comprehension tool that provides insights from individual WGS research articles, and (iv) an AI inverse modelling tool that employs a theory-guided AI model to identify suitable catalyst designs. 

AceWGS is built using open-source software and moderate-sized LLMs (e.g., Llama3 with 8 billion parameters), ensuring adaptability and ease of implementation. This design enables researchers to follow the methods outlined in this article to develop their LLM-assisted frameworks. Our approach will facilitate seamless cross-disciplinary research in AI-accelerated material design.

\section{Methodology}
\label{sec:method}
\begin{figure}
    \centering
    \includegraphics[width=0.85\linewidth]{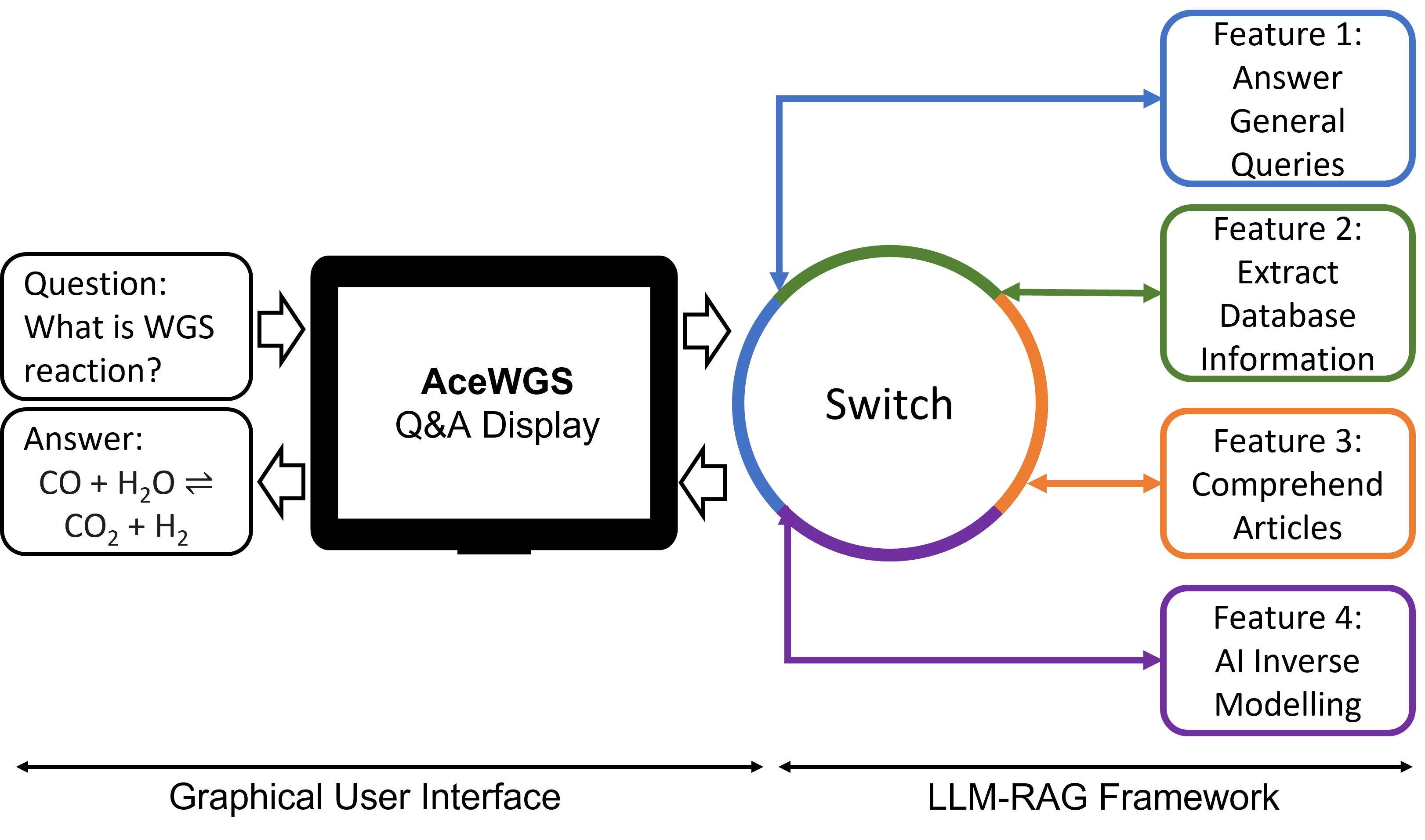}
    \caption{AceWGS framework utilizing a graphical user interface (GUI), large language models (LLMs) with retrieval augmented generation (RAG) to accelerate catalyst design for Water-Gas Shift Reactions (WGS).}
    \label{Fig:AceWGS}
\end{figure}
\subsection{AceWGS}
AceWGS consists of two primary components: a graphical user interface (GUI) and a core LLM-RAG framework, which incorporates four key features along with a central module, called "Switch" (see Fig.~\ref{Fig:AceWGS}). The module functions as a rule-based system that selects and manages which feature to activate based on researcher queries. For example, when the query specifies "inverse model", Switch configures the environment to assist researchers in identifying optimal catalyst compositions and reaction conditions. 

The entire AceWGS is built upon Python (version $3.11$)~\cite{Python}, providing the foundation for all components. The graphical user interface (GUI) is developed using Tkinter~\cite{Tkinter}, offering researchers an intuitive way to interact with the framework. LangChain (version $0.3$)~\cite{Langchain} is used to orchestrate large language models (LLMs) and local databases, enabling seamless retrieval-augmented generation (RAG) processes. Ollama (version $0.3.11$)~\cite{Ollama} manages the execution of LLMs, with open-source models such as Llama installed locally on a Ubuntu machine, facilitating efficient and flexible model deployment for the LLM-RAG framework.

The following sections describe the methodologies behind each of the four features: answering general queries, extracting database information, comprehending research articles, and AI inverse modelling.

\subsection{Answer General Queries}
\begin{figure}
    \centering
    \includegraphics[width=0.8\linewidth]{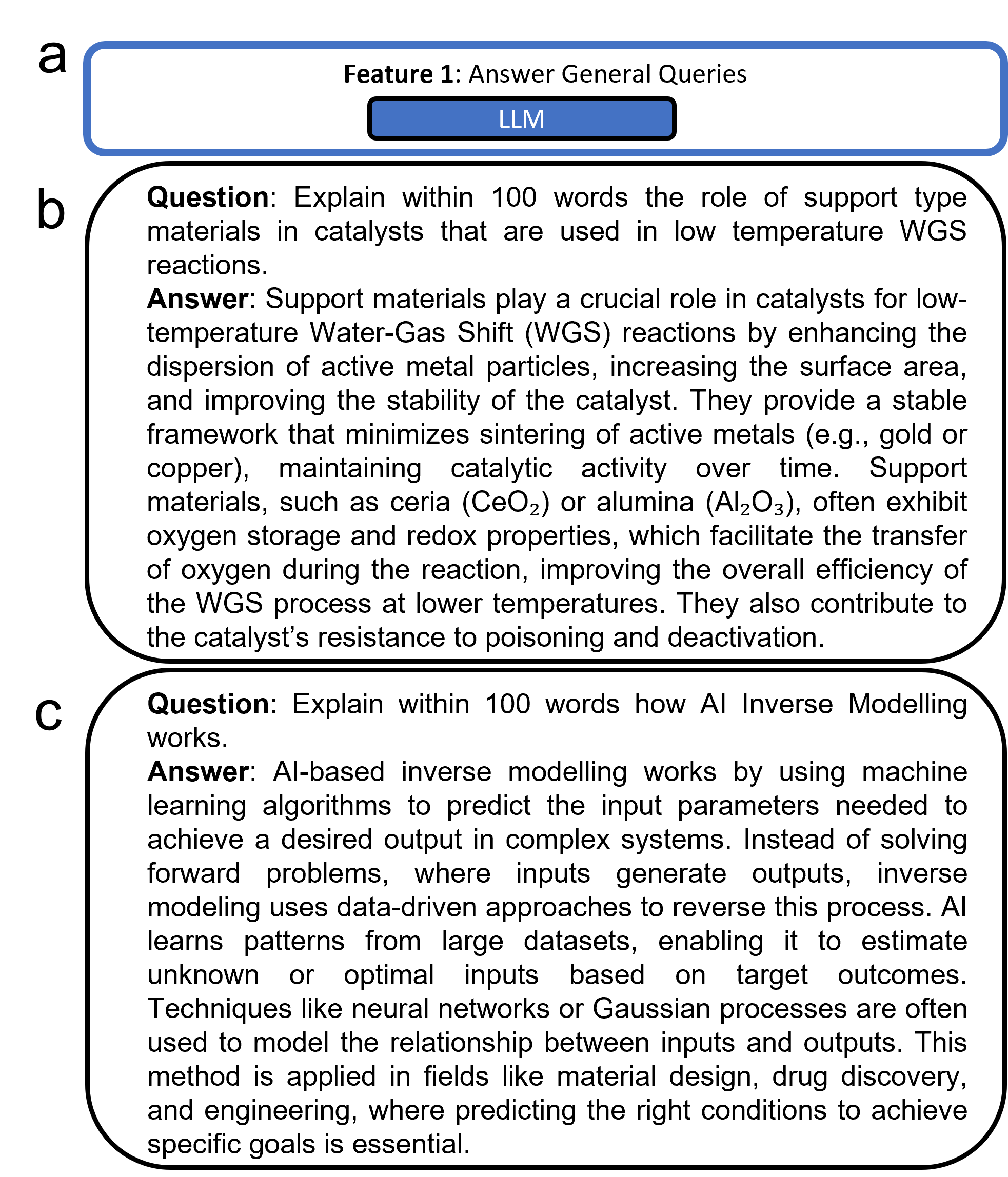}
    \caption{(a) Feature-1 contains an LLM model to answer general queries. (b) A typical answer generated by Feature-1 is based on a catalyst-related question. (c) Similarly, a typical answer generated by Feature-1 is based on an AI-related question.}
    \label{Fig:general}
\end{figure}
The "Answer General Queries" feature is designed to answer various general inquiries, including WGS reactions, catalysts, machine learning, inverse modelling, generative AI, LLMs, and RAGs. Leveraging open-source LLM models, this feature delivers domain-specific responses with high accuracy. Its primary purpose is to facilitate interdisciplinary communication, allowing researchers from different fields to access relevant information from outside their areas of expertise quickly. The framework is straightforward, utilizing an LLM, with typical question-and-answer examples shown in Fig.~\ref{Fig:general}.

\subsection{Extract Database Information}
\begin{figure}
    \centering
    \includegraphics[width=0.85\linewidth]{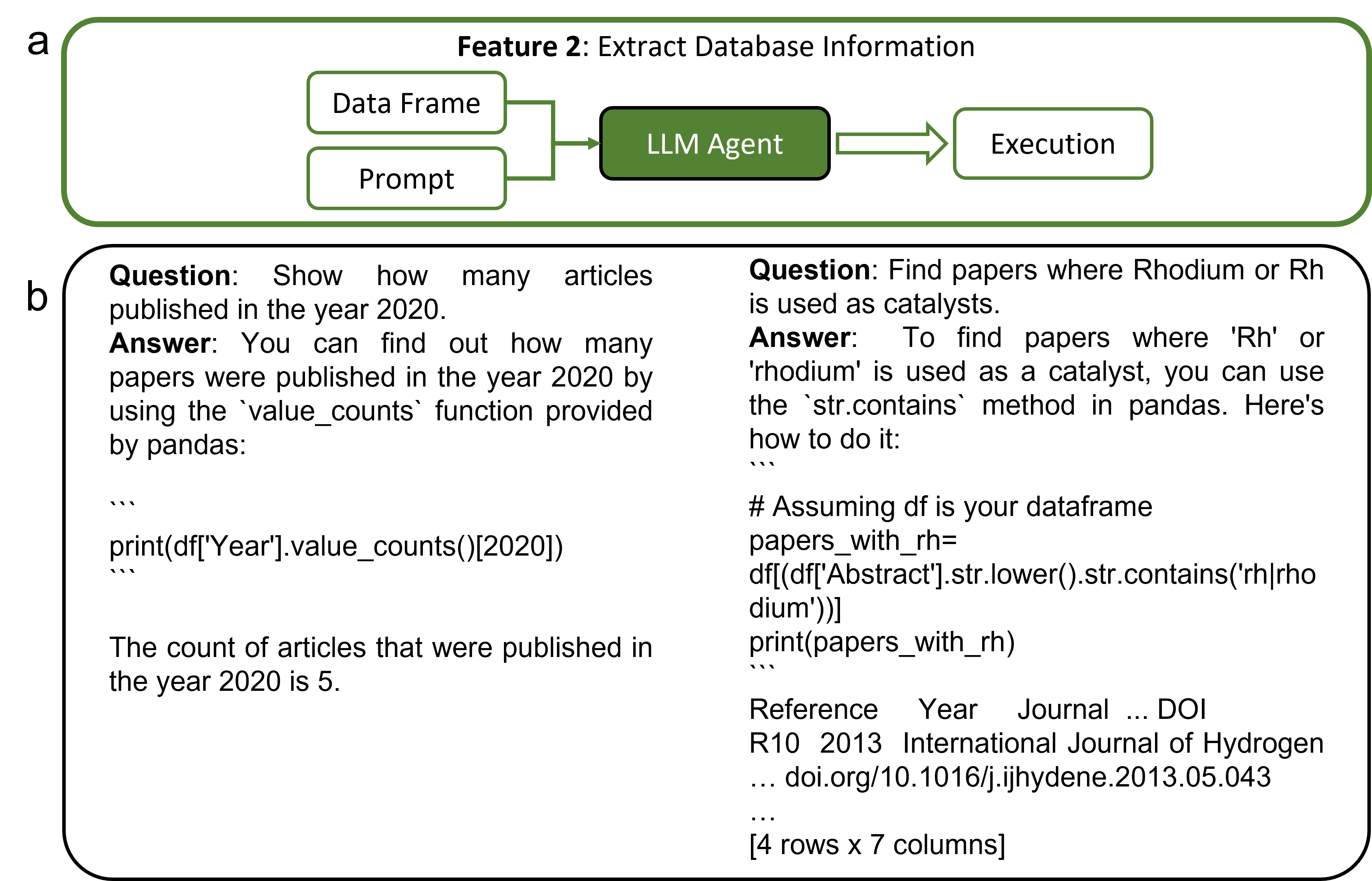}
    \caption{(a) Feature-2 consists of two tools: (i) an LLM agent that takes a data frame and a customized prompt as inputs, and (ii) an execution tool that runs the Python command suggested by the LLM agent. (b) Typical responses generated by Feature-2 in answer to two questions designed to retrieve information from the local database.}
    \label{Fig:extract}
\end{figure}
The "Extract Database Information" feature helps researchers retrieve qualitative, quantitative, and statistical information from a local database of 82 articles on WGS experiments involving noble metal catalysts. To achieve the objective, we first constructed a data frame using pandas library~\cite{pandas}, which contains seven metadata fields for each article: reference ID, publication year, title, abstract, journal name, author names, and digital object identifier (DOI). We then formulated a prompt describing the data frame and its seven metadata fields. The data frame and the prompt were inputted into an inbuilt LLM agent, {\it create\_pandas\_dataframe\_agent}~\cite{dfAgent}, within LangChain, which is designed to interpret data frames and perform operations such as data retrieval, and filtering, based on customized queries. The agent generates suggested Python commands, which are then passed to an execution tool that runs these commands and displays the corresponding results (see Fig.~\ref{Fig:extract}(a)). 

Fig.~\ref{Fig:extract}(b) presents two sample queries and corresponding answers about the local database, demonstrating the utility of this feature in efficiently extracting insights. By retrieving and organizing relevant data, this tool aids researchers in evaluating the strengths and limitations of the database, facilitating informed decision-making for further analysis.

\subsection{Comprehend Articles}
\begin{figure}
    \centering
    \includegraphics[width=0.85\linewidth]{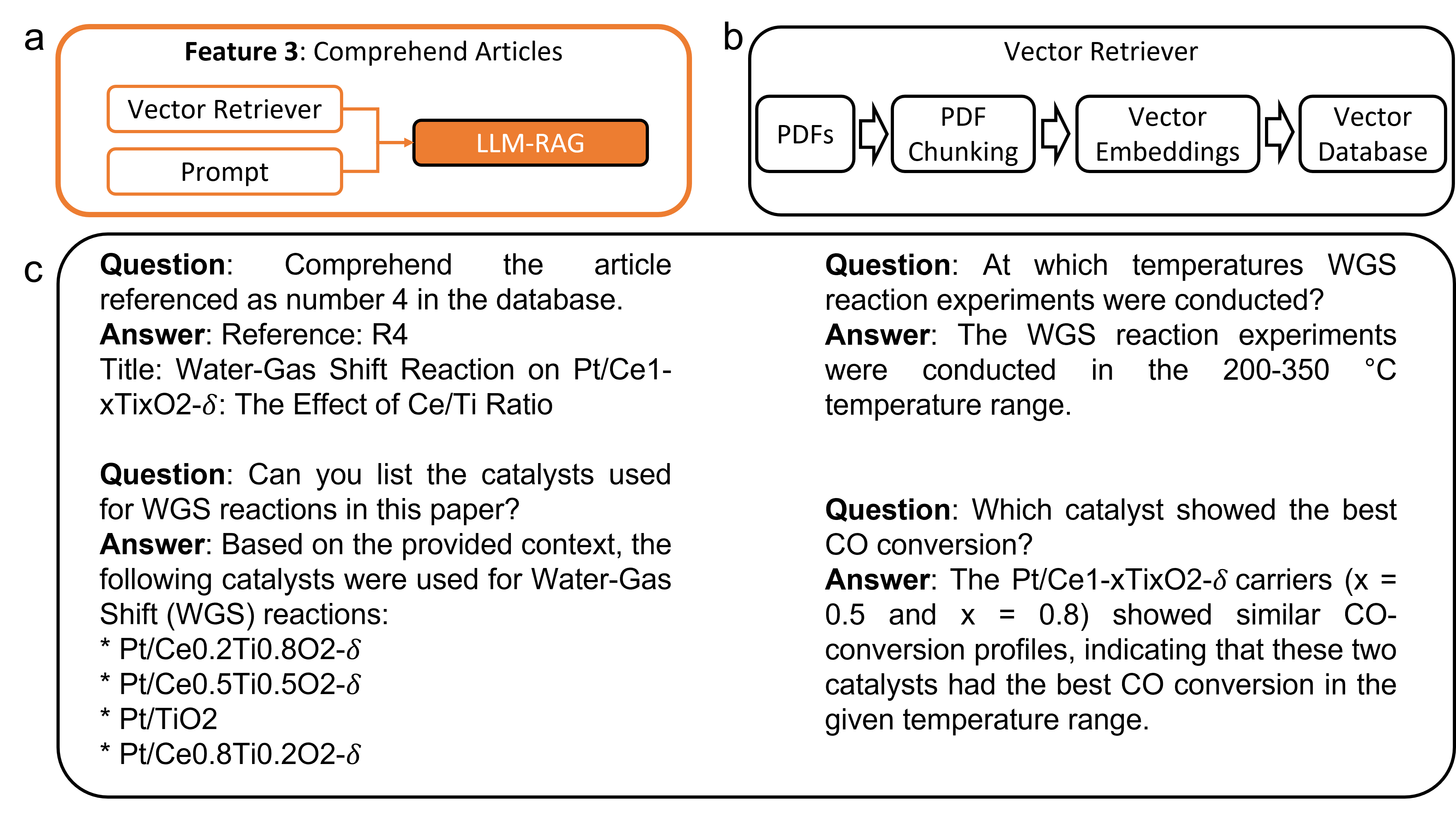}
    \caption{(a) Feature-3 is an LLM-RAG that takes a vector retriever tool and a customized prompt as inputs. (b) The workflow of the vector retriever tool. (c) Typical responses generated by Feature-3 in answer to four questions set to comprehend a research article.}
    \label{Fig:comprehend}
\end{figure}
The "Comprehend Articles" feature allows researchers to retrieve information from any of the 82 articles stored in the local database. To use the feature, researchers first input the reference ID of the desired article into the GUI, and then proceed to pose questions. The system processes each query using an LLM-RAG framework, which is integrated with a customized prompt and a vector retriever tool (see Fig.~\ref{Fig:comprehend}(a)). The prompt guides the LLM-RAG system in identifying the comprehension task and generating an appropriate response. The vector retriever tool allows the system to search and retrieve the most relevant sections of the article efficiently, ensuring that the LLM can accurately focus on and answer specific questions by accessing the semantically related portions of the document.

The workflow of the vector retriever tool is illustrated in Fig.~\ref{Fig:comprehend}(b). The process begins with text extraction (often referred to as "chunking") from 82 articles from their portable document format (PDF). 
Each PDF is segmented into a list of 1000-character texts with a 150-character overlap between adjacent segments using pdfMiner~\cite{PDFMiner} and {\it RecursiveCharacterTextSplitter} module of LangChain.
 
 Next, the extracted list of text segments is converted into numbers, specifically vector embeddings, for further processing by an LLM. The vectorization is executed using the {\it mxbai-embed-large} model of OllamaEmbeddings~\cite{OllamaEmbeddings}, and the resulting embeddings are stored in a vector database built with FAISS~\cite{FAISS}.

We demonstrated an example of the feature in Fig.~\ref{Fig:comprehend}(c), where we selected an article from our database, identified as reference 4. The generated answers provided insights into several key aspects, including the number and types of catalysts, the reaction temperatures for conducting WGS experiments, and the corresponding results. Specifically, in this case, the feature enabled the identification of the best-performing catalysts.

\subsection{AI Inverse Modelling}
\begin{figure}
    \centering
    \includegraphics[width=0.85\linewidth]{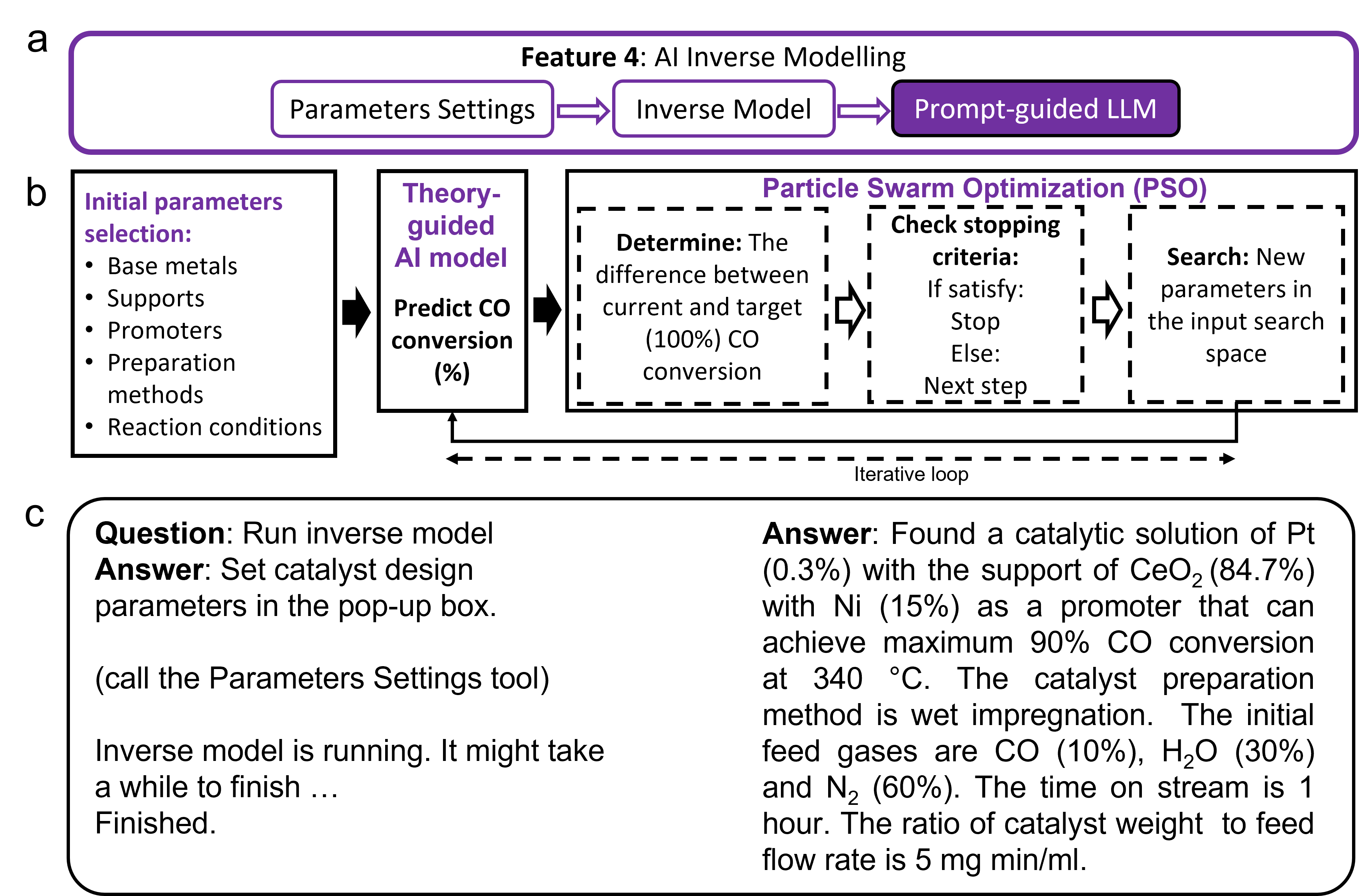}
    \caption{(a) Feature-4 contains three tools: (i) the Parameter Settings tool, a GUI where researchers can set the required catalyst design parameters, e.g., base metals, supports, promoters, preparation methods, and reaction conditions; (ii) the Inverse Model, which searches for the best catalytic candidates based on the set design parameters; and (iii) the Prompt-guided LLM, which takes the outputs of the Inverse Model and explains them in a natural language manner. (b) The Inverse Model framework, where the model initially takes a set of parameters, predicts the CO conversion percentage using our theory-guided AI model then performs particle swarm optimization to search for catalytic candidates. (c) A typical query on inverse modelling results in a sequence of responses starting from setting parameters, stating the status of inverse modelling, and finally displaying the solution.
    }
    \label{Fig:inverse}
\end{figure}
The "AI Inverse Modelling" feature searches for optimal catalytic candidates and reaction conditions according to the design constraints specified by researchers. This feature consists of three tools: (i) the Parameter Settings tool, a GUI where researchers can design a catalyst by selecting base metals, supports, promoters, and catalyst preparation methods, and can set reaction conditions; (ii) the Inverse Model, which identifies the best catalytic candidates based on these parameters; and (iii) the Prompt-guided LLM, which interprets and explains the outputs of the Inverse Model tool in natural language (see Fig.~\ref{Fig:inverse}(a)).

The Inverse Model tool is sophisticated, integrating our previously proposed pre-trained theory-guided AI model with a particle swarm optimization algorithm (see Fig.~\ref{Fig:inverse}(b)). The theory-guided AI model accepts catalytic compositions (numerical variables), preparation methods (binary features), and reaction conditions (numerical variables) as inputs. It outputs the predicted CO conversion percentage, the uncertainty in this prediction, and the thermodynamic equilibrium conversion for CO. The particle swarm optimization algorithm further explores and exploits these predictions to iteratively identify the optimal solution. Detailed descriptions and source codes for the model are provided in~\cite{Chattoraj2023}.

The output of the Inverse Model tool is passed to a prompt-guided LLM as a string. The LLM receives specific instructions via a prompt on how to interpret the string and then generates a concise explanation, simplifying the content into natural language in no more than 200 words. Translating the raw output string into natural language allows researchers to easily understand complex data and results, making the information more accessible for their work. 

A typical example of the AI Inverse Modelling feature is shown in Fig.~\ref{Fig:inverse}(c), where on the Parameter Settings we design a catalyst with platinum (Pt) as a base metal, cerium oxide (CeO$_2$) as support, nickel (Ni) as promoter, wet impregnation as the preparation method, and set the range of reaction temperature between 300 and 350~$^\circ$C. 
The Parameters Settings passed these initial parameters to the Inverse Model, which then found the catalyst weight percentage and optimal reaction conditions that can achieve the highest CO conversion. 
The final catalytic parameters were passed to the prompt-guided LLM which then expressed the raw parameters into natural language.

\section{Results}
\begin{table}
    \centering
    \begin{tabular}{|c|c|c|c|}
        \hline
         Model    &  General Queries &  Extraction & Comprehension \\
         \hline
         Llama2   &      2.9            &  3.0 &  3.17\\
         \hline
         Llama3   &      3.6            &  3.8 &  3.17\\
         \hline
         Llama3.1 &      4.3            &  4.1 &  3.08\\
         \hline
         Gemma2   &      4.6            &  4.3 &  3.50\\
         \hline
    \end{tabular}
    \caption{The performance of the four LLMs is presented as the average Likert score across all questions, with a maximum possible score of 5 and a minimum possible score of 1.}
    \label{Table:results}
\end{table}
In this study, we evaluate the performance of four moderately sized open-source large language models (LLMs), including Llama2 (7 billion parameters, 3.8 GB), Llama3 (8 billion parameters, 4.7 GB), Llama3.1 (8 billion parameters, 4.7 GB), and Gemma2 (9 billion parameters, 5.4 GB), across three distinct tasks of AceWGS: Answer General Queries, Extract Database Information, and Comprehend Articles. Note that we did not assess the fourth task, AI Inverse Modelling, as the role of a prompt-guided LLM in this feature is primarily auxiliary. 
We configured the LLMs with the following default parameters: ${\it temperature} = 0$, ${\it top\_k} = 10$, and ${\it top\_p} = 0.5$, to ensure that the generated responses prioritize accuracy and factual correctness over creativity, as maintaining the integrity of scientific information was the primary objective.
 
To evaluate the performance of the four LLMs for the feature Answer General Queries, we posed 10 questions, consisting of five on WGS reactions and noble metal catalysts, one on AI, one on inverse modelling, and three on LLM and RAG. Each question was evaluated using a 5-point Likert scale, where the criteria included: 1 (incorrect), 2 (poor), 3 (acceptable), 4 (good), and 5 (very good). Given the descriptive nature of the questions, there is no single correct answer, however, the answer can be potentially evaluated as incorrect. We found that Llama2 had the lowest average score of 2.9 across the ten questions, followed by Llama3, Llama3.1, and Gemma2 (see Table~\ref{Table:results}).

Similarly, for the feature, Extract Database Information, we prepared 10 questions and evaluated the performance of the LLMs using the 5-point Likert scale. In this case, each question had a single accurate answer, which could receive a score of 5, while incorrect answers were scored as 1. The evaluation revealed a performance improvement in the following order: LLama2, Llama3, Llama3.1, and Gemma2, respectively (see Table~\ref{Table:results}).  

The feature, Comprehend Articles, is an LLM-RAG system, and its effectiveness depends on several critical tasks, including accurate text segmentation, vector embeddings, retrieval search, and generation. This study focuses on the performance of the four LLMs primarily responsible for the generation task. A set of 12 questions was designed, with 4 questions each for information retrieval from three selected articles~\cite{petallidou2013water,runxia2016cu,cai2021hydrogen}. Most of these questions allow for broad but precise answers, which were evaluated using a Likert scale ranging from 1 to 5. 
Unlike the trends observed in the previous two features, the performance of the three Llama models did not improve with the newer versions for the comprehension task. However, Gemma2 consistently outperformed the other models, maintaining its superiority across various features
(see Table~\ref{Table:results}).

\section{Case study}
In this section, we demonstrate the practical utility of the AceWGS framework by presenting a case study centered on catalyst design for the WGS reaction. Specifically, we aim to identify a noble metal catalyst capable of achieving maximum CO conversion at temperatures below 300~$^\circ$C.

\subsection*{Extract Database Information}
First, identifying potential catalytic candidates reported in the literature is essential. In this scenario, we aim to find the most recent articles available in our local WGS database. This can be accomplished using the "Extract Database Information" feature of AceWGS. The typical flow of questions and corresponding answers from this feature is shown below, illustrating how researchers can efficiently gather relevant data.\\ 

\makebox[\textwidth][c]{%
    \fbox{%
        \parbox{0.9\textwidth}{%
        \textbf{Question:} Extract the journal names for all papers that were published in the year 2021. \\ 
        \textbf{Answer:}  ['Nature', 'Energy \& Fuels', 'Nanomaterials', 'Catalysis Today', 'Journal of Catalysis', 'Journal of Catalysis', 'Catalysts', 'Heliyon', 'International Journal of Energy Research', 'Catalysts'] \\
        \textbf{Question:} Retrieve the reference and title of all papers published in the year 2021 in the journal Nature.\\
        \textbf{Answer:} R71 \ 2021 \ Nature \ A stable low-temperature H2-production catalyst by crowding Pt on $\alpha$-MoC. \\
        \textbf{Question:} Retrieve papers where the string 'MoC' is mentioned in the abstract in the exact same form.\\
        \textbf{Answer:} \\
        \ \ R51 \ 2017 \ Science \  Atomic\-layered Au clusters on $\alpha$-MoC as catalysts for the low-temperature water-gas shift reaction \\
        \ \ R71 \ 2021 \ Nature \ A stable low-temperature H2-production catalyst by crowding Pt on $\alpha$-MoC 
        }
    }
} \\
\\

At this stage, we identified two articles~\cite{yao2017atomic, zhang2021stable} that report gold and platinum-based catalysts supported by $\alpha$-molybdenum carbide. From their abstracts, it is clear that both catalysts show superior performance at the desired low-temperature limit, indicating that these studies warrant further exploration.

\subsection*{Comprehend Articles}
To gain a deeper understanding of the two identified articles, we employed the Comprehend Articles feature of AceWGS. The typical workflow of questions and their corresponding answers is illustrated below. \\

\makebox[\textwidth][c]{%
    \fbox{%
        \parbox{0.9\textwidth}{%
         \textbf{Question:} Comprehend the article of reference ID R71.\\
         \textbf{Answer:} Ready to retrieve information from the article R71. \\  
         Title: A stable low-temperature H$_2$-production catalyst by crowding Pt on $\alpha$-MoC \\
         \textbf{Question:} Extract the name of the catalysts mentioned in the article. \\
         \textbf{Answer:} The catalysts mentioned in the article are Pt/{$\alpha$}-MoC catalysts with varying Pt loadings.\\  
         \textbf{Question:} Find the name of the catalyst synthesis or preparation method. \\
         \textbf{Answer:} The catalyst synthesis method used is incipient wetness impregnation (IWI).  \\
         \textbf{Question:} Provide a step-by-step synthesis method for the catalyst as described in the article.  \\
         \textbf{Answer:} The synthesis of Pt/$\alpha$-MoC catalysts involves several steps: \\
        1. Synthesis of $\alpha$-MoC: Ammonium molybdate tetrahydrate is calcined to form MoO$_3$, which is then heated in ammonia and subsequently in a methane/hydrogen mixture to produce $\alpha$-MoC.\\
        2. Reduction of $\alpha$-MoC: The synthesized $\alpha$-MoC is reduced in a hydrogen/nitrogen mixture at 523 K for 60 minutes to create surface anchoring sites for platinum. \\
        3. Impregnation with Pt Precursor: A solution of H$_2$PtCl$_6$-6H$_2$O is mixed with the reduced $\alpha$-MoC powder, followed by drying in a vacuum oven. \\
        4. Reduction and Activation: The impregnated sample is mildly reduced at 623 K for 1 hour under hydrogen flow. Then, it's exposed to a methane/hydrogen mixture and heated to 863 K for 2 hours to reactivate the catalyst. 
    }
    }   
} \\
\\

Accordingly, by employing the Comprehend Articles feature, we gathered essential information regarding reaction conditions, catalytic performance for CO conversion, and other significant details from both studies identified through the data extraction process. With these insights, we now seek to identify gaps in the existing literature that will further guide us toward future research directions.      

\subsection*{Answer General Queries}
To facilitate research gap identification and idea generation, we pose the following question to the feature Answer General Queries of AceWGS: \\

\makebox[\textwidth][c]{%
    \fbox{%
        \parbox{0.9\textwidth}{%
         \textbf{Question:} Provide one or two catalyst design ideas based on the two existing catalysts: i. Pt $\alpha$-MoC catalyst and ii. Au $\alpha$-MoC catalyst. \\
         \textbf{Answer:} Here are a couple of catalyst design ideas: \\
         1. Alloying:  Combine Pt and Au in an $\alpha$-MoC support to potentially leverage the high activity of Pt at lower temperatures and the stability of Au at higher temperatures. This could create a catalyst with a broader operating temperature range.\\ 
         2. Tuning Particle Size Distribution: Explore varying the size distribution of Pt or Au nanoparticles on the $\alpha$-MoC support.  The text highlights that different sizes (isolated atoms, clusters, particles) can influence activity and stability. Optimizing this distribution might lead to enhanced performance. 
   }
    }   
} \\
\\

Both ideas presented above are quite compelling. We further investigate the first concept of "Alloying" by utilizing our AI Inverse Modelling feature.

\subsection*{AI Inverse Modelling}
Once we input platinum, gold, and $\alpha$-molybdenum carbide as the catalyst combination, select incipient wetness impregnation (IWI) as the preparation method, and set the reaction temperature below 300~$^\circ$C, the AI Inverse Modelling feature of AceWGS generates the following catalytic solution:\\

\makebox[\textwidth][c]{%
    \fbox{%
        \parbox{0.9\textwidth}{%
         \textbf{Question:} Run inverse model. \\
         \textbf{Answer:} Set catalyst design parameters in the pop-up box. \\
         (call the Parameters Settings tool)\\
         The inverse model is running. It might take a while to finish ... \\
         Finished. \\
         \textbf{Answer:} Found a catalytic solution of Pt (4.26\%), Au (3.09\%) with the support of $\alpha$-MoC (92.64\%) that can achieve maximum 95.07\% (error $\pm$ 0.79\%) CO conversion at 200~$^\circ$C. The catalyst preparation method is incipient wetness impregnation (IWI). The initial feed gases are CO (0.1\%), H$_2$O (6.18\%), CO$_2$ (5\%), H$_2$ (0.15\%), and N$_2$ (88.57\%). The time on stream is 1 hour. The ratio of catalyst weight to feed flow rate is 1 mg min/ml. 
   }
    }   
} \\
\\

The generated results now provide a comprehensive catalyst formulation, detailing the exact weight percentages for each catalyst component, the specific volume ratios of initial feed gases, and optimized reaction conditions. In addition to identifying the preparation method (IWI), AceWGS supplies an in-depth synthesis protocol, as retrieved through the Comprehend Articles feature. This level of detail enhances reproducibility and provides critical insights for further experimentation and validation.

\section{Conclusion}
In conclusion, we present AceWGS, a versatile LLM-RAG framework designed to streamline AI-driven catalyst discovery for the water-gas shift (WGS) reaction. AceWGS provides four key features that enhance cross-disciplinary collaboration, enable efficient retrieval of state-of-the-art information, and optimize textual and numerical data extraction from literature, ultimately identifying promising catalytic candidates in a fraction of the time required by traditional methods. Purposefully constructed with open-source tools and moderate-sized LLMs, AceWGS establishes a prototype accessible to researchers with standard computational resources. It allows them to adapt and extend the framework for accelerated, cross-domain research in AI-powered materials design.

In the future, We will focus on expanding AceWGS by integrating advanced features, automating data retrieval directly from the literature, and streamlining dataset preparation for AI model training. These enhancements aim to improve efficiency and support more sophisticated AI-driven workflows for catalyst design.

\section*{Acknowledgement}
This research is supported by Agency for Science Technology and Research (A*STAR) $\langle$C222812014$\rangle$.



\bibliographystyle{elsarticle-num} 

\end{document}